\documentclass[
reprint
]{revtex4-2}

\usepackage{amsmath}
\usepackage{amsfonts}
\usepackage{amssymb}
\usepackage{graphicx}
\usepackage{hyperref}
\usepackage{tikz}
\usepackage{pgfplots}
\usepackage{booktabs}
\usepackage{cleveref}
\pgfplotsset{width=\linewidth,compat=1.9}

\usepackage{natbib}
\begin{document}

\title{Generative time series models using Neural ODE in Variational Autoencoders}

\author{Marcus Lenler Garsdal, s174440}
\author{Valdemar Søgaard, s174431}
\author{Simon Moe Sørensen, s174420}

\affiliation{Technical University of Denmark (DTU)\\
	02456 Deep Learning Fall 21}

\date{\today}
\begin{abstract}

In this paper we present an extension to the work done in \cite{chen2019a} by implementing Neural Ordinary Differential Equations in a Variational Autoencoder setting for generative time series modeling. An object-oriented approach to the code was taken to allow for easier development and research and all code used in the paper can be found in this \href{https://github.com/simonmoesorensen/neural-ode-project}{Github link}.

The results in \cite{chen2019a} were initially recreated and the reconstructions compared to a baseline Long-Short Term Memory AutoEncoder. The model was then extended with a LSTM encoder and challenged by more complex data consisting of time series in the form of spring oscillations. The model showed promise, and was able to reconstruct true trajectories for all complexities of data with a smaller RMSE than the baseline model. However, it was able to capture the dynamic behavior of the time series for known data in the decoder but was not able to produce extrapolations following the true trajectory very well for any of the complexities of spring data. A final experiment was carried out where the model was also presented with 68 days of solar power production data, and was able to reconstruct just as well as the baseline, even when very little data is available. 

Finally, the models training time was compared to the baseline. It was found that for small amounts of data the NODE method was significantly slower at training than the baseline, while for larger amounts of data the NODE method would be equal or faster at training.

The paper is ended with a future work section which describes the many natural extensions to the work presented in this paper, with examples being investigating further the importance of input data, including extrapolation in the baseline model or testing more specific model setups.

\end{abstract}
\maketitle
\section{Introduction}


In the recent years, deep learning has shown great advances due to increased computational power and availability of different frameworks. As an example, it has shown exceptional ability in image recognition tasks using the ResNet framework. In 2019, as an extension based on ResNet, a new framework was introduced which uses Ordinary Differential Equations (ODEs) to represent hidden layers in continuous time rather than discrete time, this type of neural net is named a NeuralODE (NODE) \cite{chen2019a}. Taking the NODE one step further, it can be utilized in continuous time extra- and interpolation for sequential data, by implementing the NODE as a decoder of a Variational Auto Encoder. This paper will investigate the robustness and possibilities of this technique.

\subsection{From ResNet to ODE's}
ResNet utilizes 'residual blocks' as the core part of its architecture. These blocks are defined at a finite and discrete number of times and the output of each block can be expressed as a function
\begin{equation}
 \mathbf{h}_{t+1}= \mathbf{h}_t + f(\mathbf{h}_{t},\theta_t)
\end{equation}
With  $t$ belonging to a finite set ranging from 0 to a fixed amount of steps $T$, corresponding to the number of blocks in the net. Going from this discrete setting to a continuous one can be done by looking at what happens when the step size becomes smaller. When the step size becomes small enough, at the limit when $t \rightarrow 0$ the change to the hidden step $\mathbf{h}_{t+1}$ can be seen as a derivative with respect to $t$ instead \cite{chen2019a}. 
\begin{equation}\label{eq:ODE}
\frac{d\mathbf{h}}{dt}= f(\mathbf{h}(t),t,\theta)
\end{equation}
This allows us to define the depth of the network in a continuous manner.
The function $f$ represents the hidden dynamics and is parameterized as a neural network. 
To solve the derivatives given in \Cref{eq:ODE} the NODE uses a black box ordinary differential equation solver, which takes as input an initial hidden state $\mathbf{h}(0)$ to solve an initial value problem up to some given time $T$. This way the ODE solver yields a representation of a continuous hidden state trajectory, instead of a discrete amount of  hidden states. This also means that any specific hidden state along the hidden trajectory can be evaluated, even with uneven step sizes, which is one of the advantages with this approach. 


\subsection{Adjoint-state method}

The training process of NODE is a bit different compared to usual Feed Neural Networks (FFNs) which needs to store the operations of the forward pass in order to compute the weights using the chain rule in a backpropagation algorithm. 

\begin{figure}[h]
    \centering
  \includegraphics[width=0.4\textwidth]{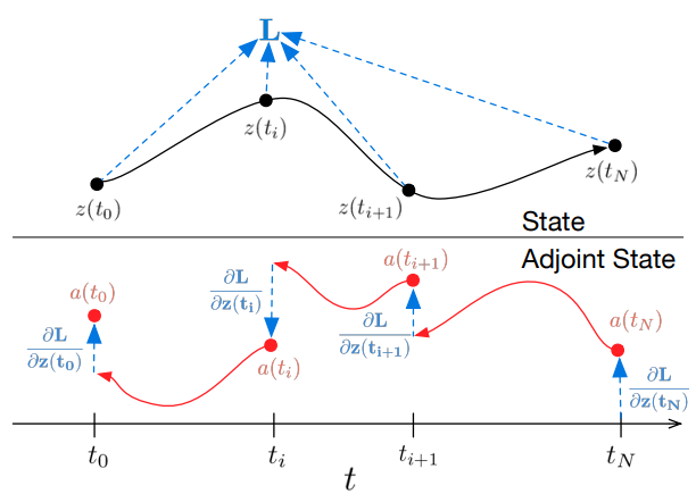}
  \caption{Reverse-mode differentation of an ODE solution\cite{chen2019a}}
  \label{fig:adjoint_state_method}
\end{figure}

In the NODE framework, no intermediate operations are stored in the forward pass (when solving the ODE of the hidden trajectory), which is what leads to the constant memory property presented in \cite{chen2019a}. Instead, when backpropagating through the network $f$ to update the weights, $\theta$, a second augmented ODE is solved, which can return the gradient of the parameters with respect to the loss function $\frac{\partial L}{\partial \theta}$. This augmented ODE is solved \textit{backwards} in time, starting from the final time step $t_1$ and going to the initial time step $t_0$. This is what allows the model not to store the operations in the forward pass. The augmented ODE takes advantage of the \textit{adjoint sensitivity method}, which is a mathematical method that can efficiently compute the gradients in an ODE. To find $\frac{\partial L}{\partial \theta}$, a quantity called the adjoint $\mathbf{a}(t)=\frac{\partial L}{\partial \mathbf{z}(t)}$ is defined, which corresponds to the gradient of the loss function with respect to the hidden state $\mathbf{z}(t)$. The derivative of the adjoint can further be expressed as 
\begin{equation}
\frac{d \mathbf{a}(t)}{d t}=-\mathbf{a}(t)^{\top} \frac{\partial f(\mathbf{z}(t), t, \theta)}{\partial \mathbf{z}}
\end{equation}
From here, the adjoint and the partial derivative of $f$ with respect to $\theta$, can be used to find the derivative of the loss function with respect to $\theta$.
\begin{equation}
\frac{d L}{d \theta}=-\int_{t_{1}}^{t_{0}} \mathbf{a}(t)^{\top} \frac{\partial f(\mathbf{z}(t), t, \theta)}{\partial \theta} d t
\end{equation}
This integral can be computed through a call to an ODE solver, which takes as input the aforementioned augmented function, comprising of the hidden trajectory $\mathbf{z}(t)$ and the adjoint state $\mathbf{a}(t)$, as well as the initial states of the augmented function (which are defined at $t_1$), namely $s_{0}=\left[\mathbf{z}\left(t_{1}\right), \frac{\partial \bar{L}}{\partial \mathbf{z}\left(t_{1}\right)}, \mathbf{0}_{|\theta|}\right]$. It is important to note that the adjoint state depends directly on the hidden trajectory $\mathbf{z}(t)$, but this can implicitly be computed, alongside the adjoint state in the call to the ODE solver. 
If the loss depends explicitly on previous states of the hidden trajectories (observations) it important to adjust the adjoint state accordingly, based on the direction of the derivative of the loss with respect to the hidden trajectory at this specific state, $\frac{\partial L}{\partial \mathbf{z}(t_i)}$. This is illustrated in \Cref{fig:adjoint_state_method}.


\subsection{Using Neural ODE's as VAE}
NODE's are intrinsically well suited for temporal data and have many advantages due to the continuous aspect of the underlying ODE. It can be used as a generative model utilizing learned representations of a latent space. In this setting it can be used to predict or extrapolate data, as well as interpolate and impute missing data within a time series. This also allows the data to be irregularly sampled when training and using the model. As a generative model, the NODE works similarly to a VAE where a latent space representation is learned through the training of the model. A VAE has an encoding and decoding part, and connecting these is a 'bottleneck' layer containing the latent space distribution. Latent states can then be sampled from the learned distribution and given to the decoder part of the network, which then reconstructs (maps) the latent states into the original space as the input data.

\begin{figure}[h]
  \includegraphics[width=0.5\textwidth]{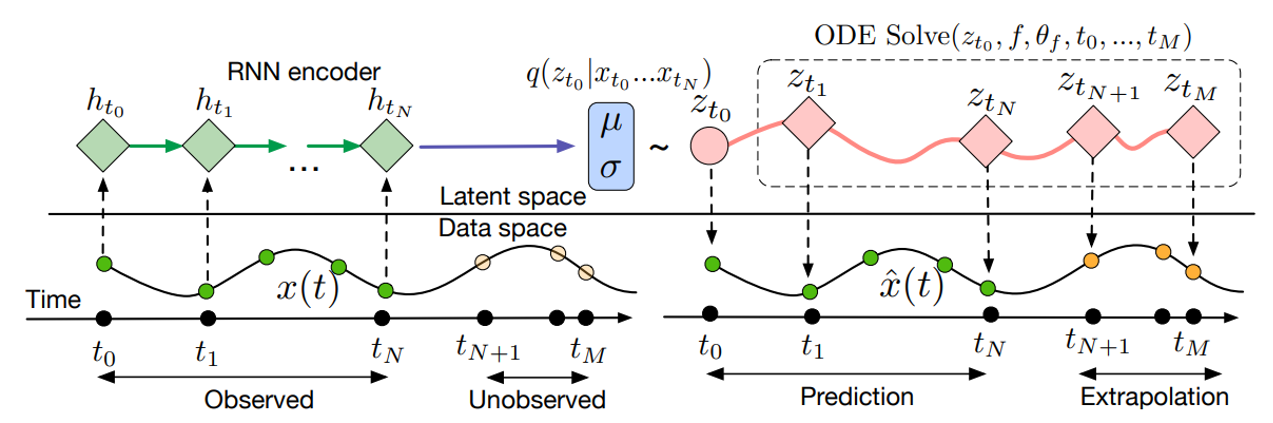}
  \caption{Structure of the neural ODE latent variable generative model\cite{chen2019a}.}
  \label{fig:neural_VAE}
\end{figure}

Our work revolves around using the NODE framework as a VAE, where it can learn the latent space distribution of a time series. In this setting the latent space corresponds to the initial state of the latent trajectory $\mathbf{z_0}$, which an ODEsolver can take as input, in order to compute the entire latent trajectory $z(t)$. The hidden trajectory is then evaluated at specific time steps, and these steps are then fed to a decoder that transforms it back into the temporal space. The workflow of the model is sketched in \Cref{fig:neural_VAE}. We will try the model on different toy examples, and on some real world solar data, and evaluate its performance against a baseline ODE to investigate the performance and limits. 

\section{Methodology and Data}
In this section we will discuss the methodology for how we have achieved the results in the paper. In addition, the data used for the experiments will be presented along with models and hyperparameters for reproducibility.


\subsection{Experiments}
We have trained the NODE model on three different data sets to test the robustness and performance of the method across multiple data sets and use cases. The first data set consists of clockwise and counterclockwise spirals, the second data set consists of three subcategories of sine waves, also referred to as \textit{springs} in the paper and the third data set consists of real world solar power curves. The data is represented as sequential data in a matrix with size: [num data, sequence, features]. The features are (x,y) coordinates for all data sets.

All data sets are split into a $60\%/20\%/20\%$ train/validation/test split. Due to different complexities, the total number of data points depend on the respective data set.
\subsubsection{Spiral}
The spiral data set is generated and consists of a total of 500 counter- and clockwise spirals, with an even 50/50 split between each, meaning there are 250 counterclockwise spirals and 250 clockwise spirals. For each spiral (data point) we have a total of 300 time steps where we train the models on a subsample of 200 time steps as seen in \Cref{fig:reconstruction_spring}. Finally, to introduce some uncertainty to the data, we introduced Gaussian noise to the subsample.

\subsubsection{Spring}
\begin{figure}[h]
  \includegraphics[width=0.5\textwidth]{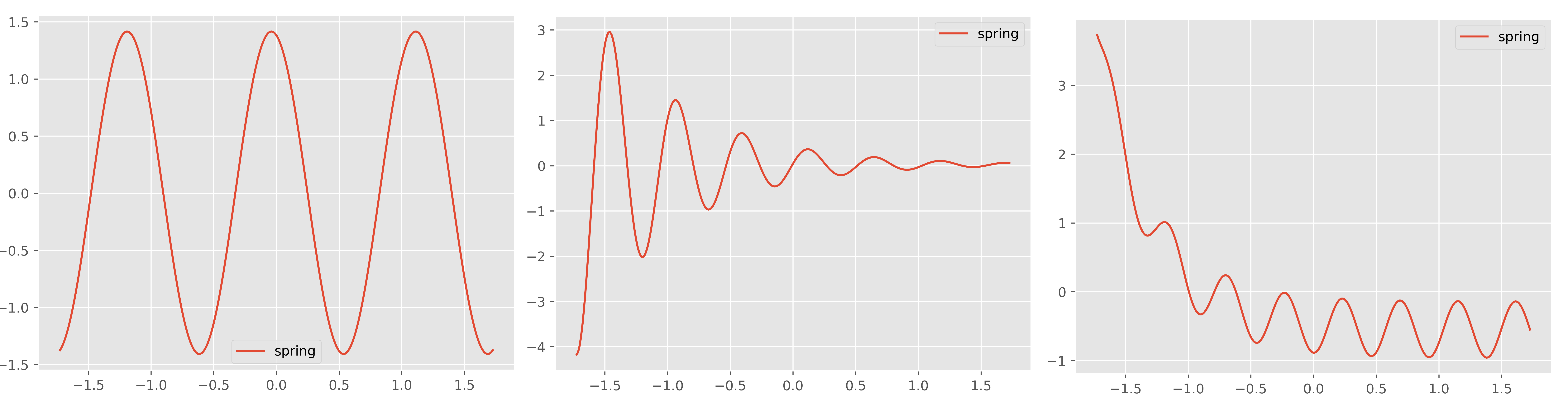}
  \caption{Different spring examples included in the data: (left) example 1, (mid) example 2 and (right) example 3.}
  \label{fig:springs}
\end{figure}

The spring data set is generated and contains three different types of springs. A non-dampened sine wave (1), a dampened sine wave (2) and an exponentially dampened sine wave (3) as seen in \Cref{fig:springs}.
The models are trained on 7 different combinations of the springs to investigate how the model performs when increasing data complexity, we refer to these combinations as \textit{mixed spring data}. The combinations are: $[1, 2, 3, 1 2, 1 3, 2 3, 1 2 3]$. For all spring data sets there are 5000 data points with 300 time steps and trained on a subsample of 200 time steps. As an uncertainty term, each new sampled oscillation would be adjusted by noise from a normal distribution which would change the amplitude and shift along the x-axis, while the frequency of the oscillation was kept constant to make it easier for the model.

\subsubsection{Solar power}
The solar power data set comes from real life solar power data where the power output of a solar cell is measured at a 30 minute interval throughout the day. This yields 48 time steps. Moreover, the total number of data points is 68 and thus not a lot of data to train on compared to the other data set.

\subsection{Code structure}
Since this paper explores a wide range of options and across multiple models and data sets, an object-oriented approach was taken to organize code, enable faster development and streamlined training. The code structure is elaborated further in the GitHub \href{https://github.com/simonmoesorensen/neural-ode-project/blob/master/README.md}{README.md}

\subsection{Models}
The VAE NODE models follow the structure seen in \cite{chen2019a} and \Cref{fig:neural_VAE}  where time series data is encoded into a latent state using a time variant neural net such as a Recurrent Neural Network (RNN), or in our case a Long-Short Term Memory (LSTM) network. The latent state is represented from $\mathbf{z}_{t_0}$ to ensure that decoding happens from the beginning of the time series. In order to enable this we encode the time series in \textit{reverse} thus ending up in $\mathbf{z}_{t_0}$. For the decoding process, an ODE layer is solved using an ODE solver over a user-defined time range to compute the latent trajectory from $\mathbf{z}(0)$ to $\mathbf{z}(t)$. This is what enabled inter- and extrapolation in continuous time. Finally this layer is passed to a standard FFNN with one hidden layer to map back into the data domain. 

For the baseline models, an LSTM Autoencoder is used to compare and evaluate the reconstructions of the NODE VAE. 

The baseline consists of a two-cell LSTM with a hidden layer of size 45, tanh activation function and 2 features in the encoded dimension. This is decoded with the model architecture but in reverse. 
The NODE VAE models are different for each data set. For the \textbf{spiral} data set the model encodes using an RNN with 45 hidden states and tanh activation function. It has 4 latent dimensions which are decoded using an ODE layer of a FFNN with 3 hidden layers, 30 hidden states and an ELU activation function. Finally, a FFNN of 1 hidden layer with 30 hidden states and ReLU activation function is used to map back into the data domain. The \textbf{spring} model encodes using a 1-layer LSTM with 60 hidden states and tanh activation function. The decoding differs from the spiral example only in 40 hidden states instead of 30. The \textbf{solar power} model has the same structure as the spring model.

The baseline model used the RMSE loss and the NODE VAE models used the ELBO loss.

It should be noted that finding the final model architectures was a highly iterative and exploratory process as good results lead to new ideas and improvements which in turn lead to better results and thus more ideas. Mostly the learning rate, model width, replacing RNN with LSTM, ODE solving method and noise in input data was configured. There is still a lot of different configurations to try out. This will be discussed in \Cref{sec:futurework}.

\subsection{Hyperparameters}
The models are trained using an ADAM optimizer with a learning rate of 0.001 for all models, and the ODE Solver utilizes the fixed-step Runge-Kutta method (rk4). The \textbf{spiral} models are trained on 10k epochs, the \textbf{spring} models are trained on 20k epochs and the \textbf{solar power} models are trained on 25k epochs.
It is also worth noting, that all training was performed using the same settings at DTU's High Performance Computing (HPC) clusters, to ensure that training time etc are comparable. 

\section{Results}\label{sec:Results}
The following section will discuss the results found from the experiments described in the previous section. 
In \Cref{sec:reconstructions_extrapolation} we compare the reconstructions and extrapolations of the NODE qualitatively, in \Cref{sec:RMSE} we analyze the RMSE values and in \Cref{sec:training_time} we analyze the training time per epoch across all models. Finally we visualize the latent space of one of the trained models in \Cref{sec:latent_space}.
\subsection{Reconstructions and extrapolation}\label{sec:reconstructions_extrapolation}
\subsubsection{Spiral} \label{sec:Toy}
\begin{figure}[h]
  \includegraphics[width=0.4\textwidth]{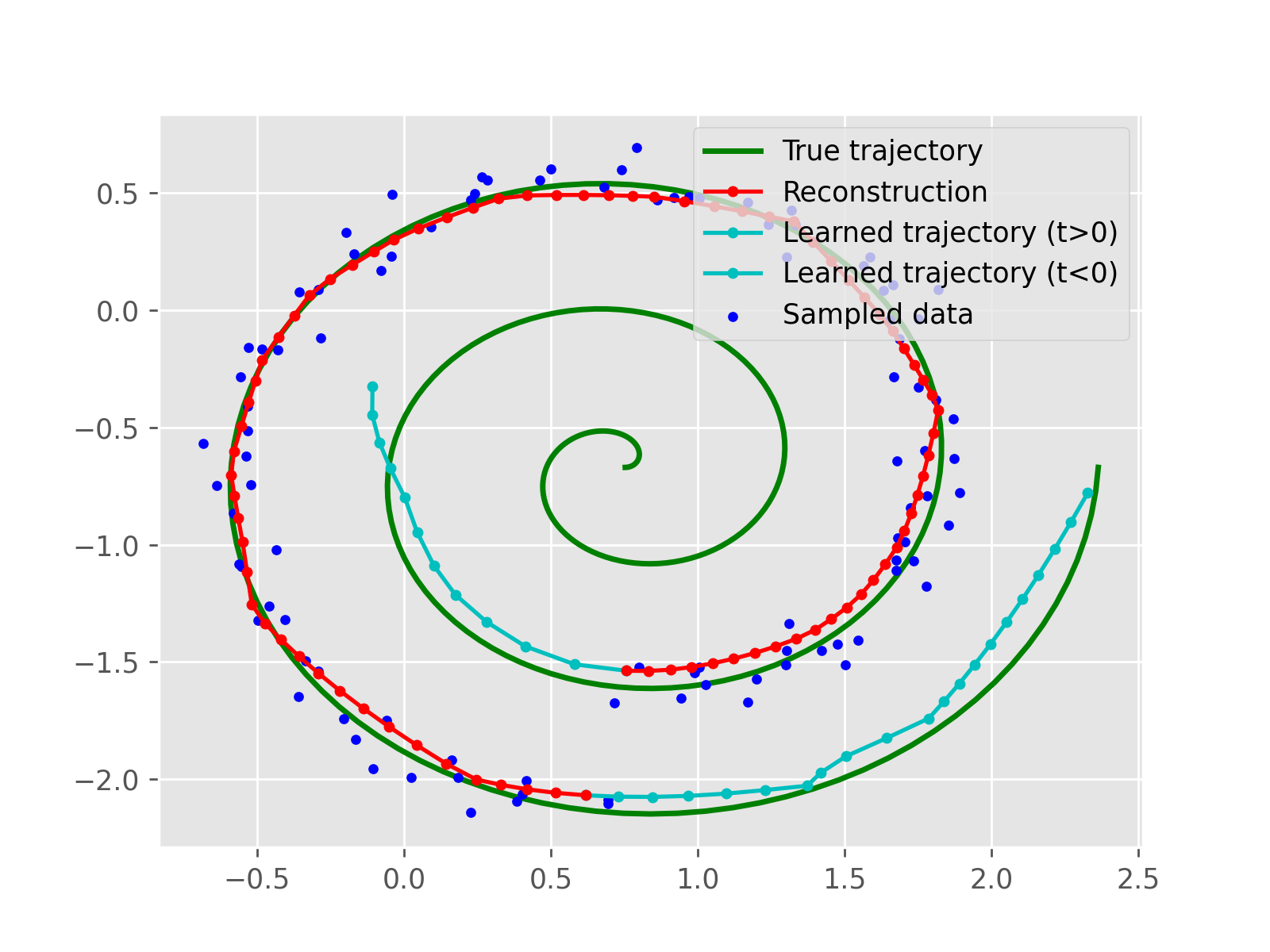}
  \caption{Reconstruction and extrapolation of a single spiral sample using NODE.}
  \label{fig:reconstruction_spiral}
\end{figure}

It can be seen from \Cref{fig:reconstruction_spiral} how the model is able to reconstruct the true trajectory when sampled data is available, as well as extrapolate for both $t<0$ and $t>0$ and follow the true trajectory. 

We have adjusted the extrapolation by changing the time $t$ presented to the ODE solver, and if $t$ is set much larger or smaller than the time $t$ which the model was trained on, then the learned trajectories begin to deviate drastically from the ground truth. Examples of this can be seen in the reconstruction grids in the appendix, e.g \Cref{fig:reconstruction_grid_spiral_node}, when both the learned backwards and forward trajectories deviate from the true trajectory. In some model runs the learned trajectory was experienced to explode to a completely different order of magnitude and never follow the true trajectory. Comparing the spiral results to \cite{chen2019a} we find that we were not able to extrapolate as far into the middle of the spiral as their results. However, we show that we can extrapolate in both directions equally well, which was not visually presented in their paper.





\subsubsection{Springs}\label{sec:Spring}

\begin{figure}[h]
  \includegraphics[width=0.45\textwidth]{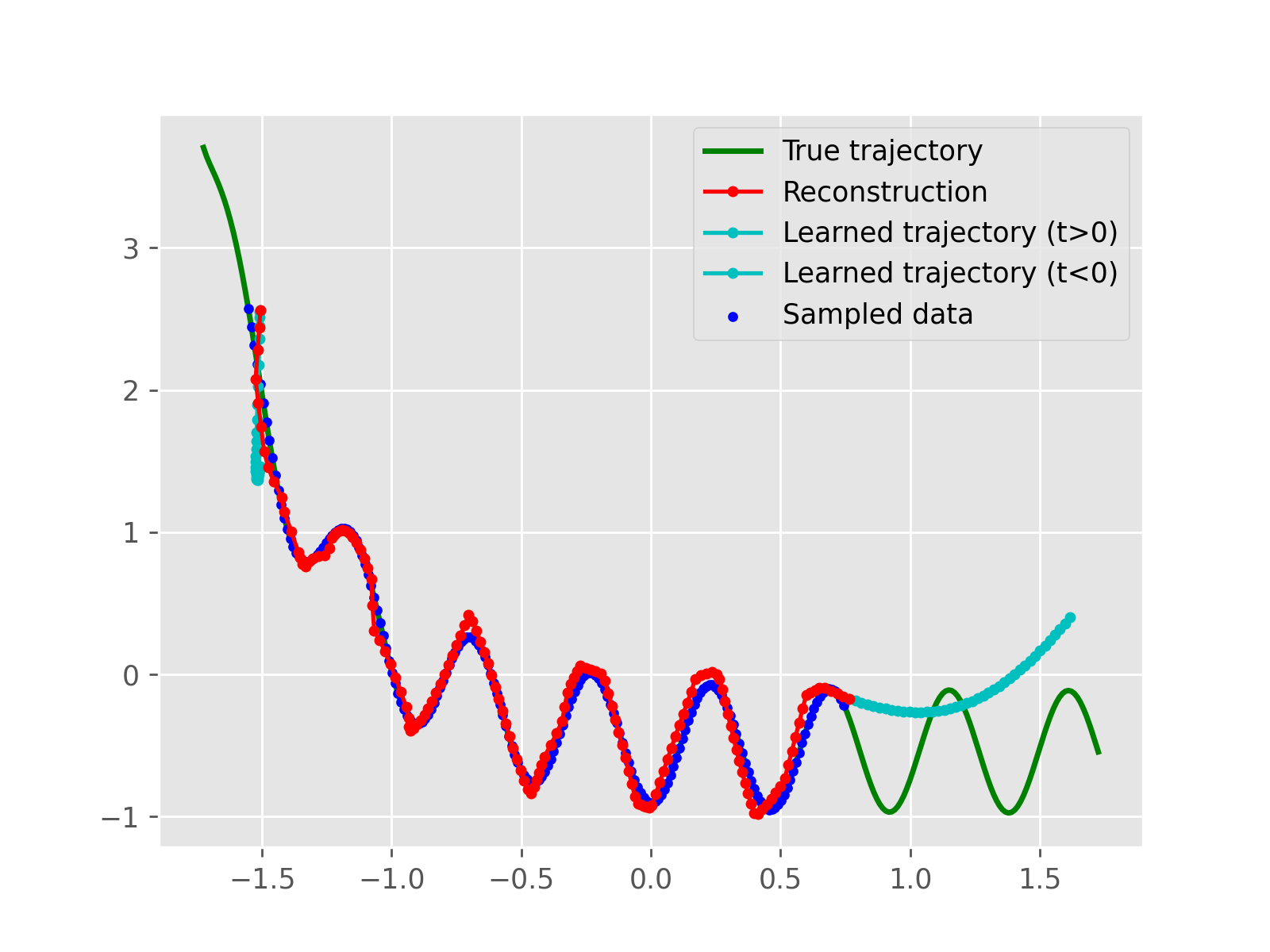}
  \caption{Reconstruction and extrapolation of a single spring sample with mixed examples 1 using NODE.}
  \label{fig:reconstruction_spring}
\end{figure}

For visualizations of the spring data set, we present the model trained on the Spring (1, 2, 3) data in \Cref{fig:reconstruction_spring}. It becomes clear that the model is able to reconstruct the original trajectory when sampled data is available, however compared to the spiral example, the extrapolation has not captured the dynamic behavior of the oscillation for both $t<0$ and $t>0$. In the appendix in \Cref{fig:reconstruction_grid_spring_node} it can be seen how for the simpler dynamics of example 1 the extrapolation follows the initial direction of the oscillation, but explodes for larger values of $t$.

\subsubsection{Solar power}\label{sec:Real}
\begin{figure}[h]
  \includegraphics[width=0.45\textwidth]{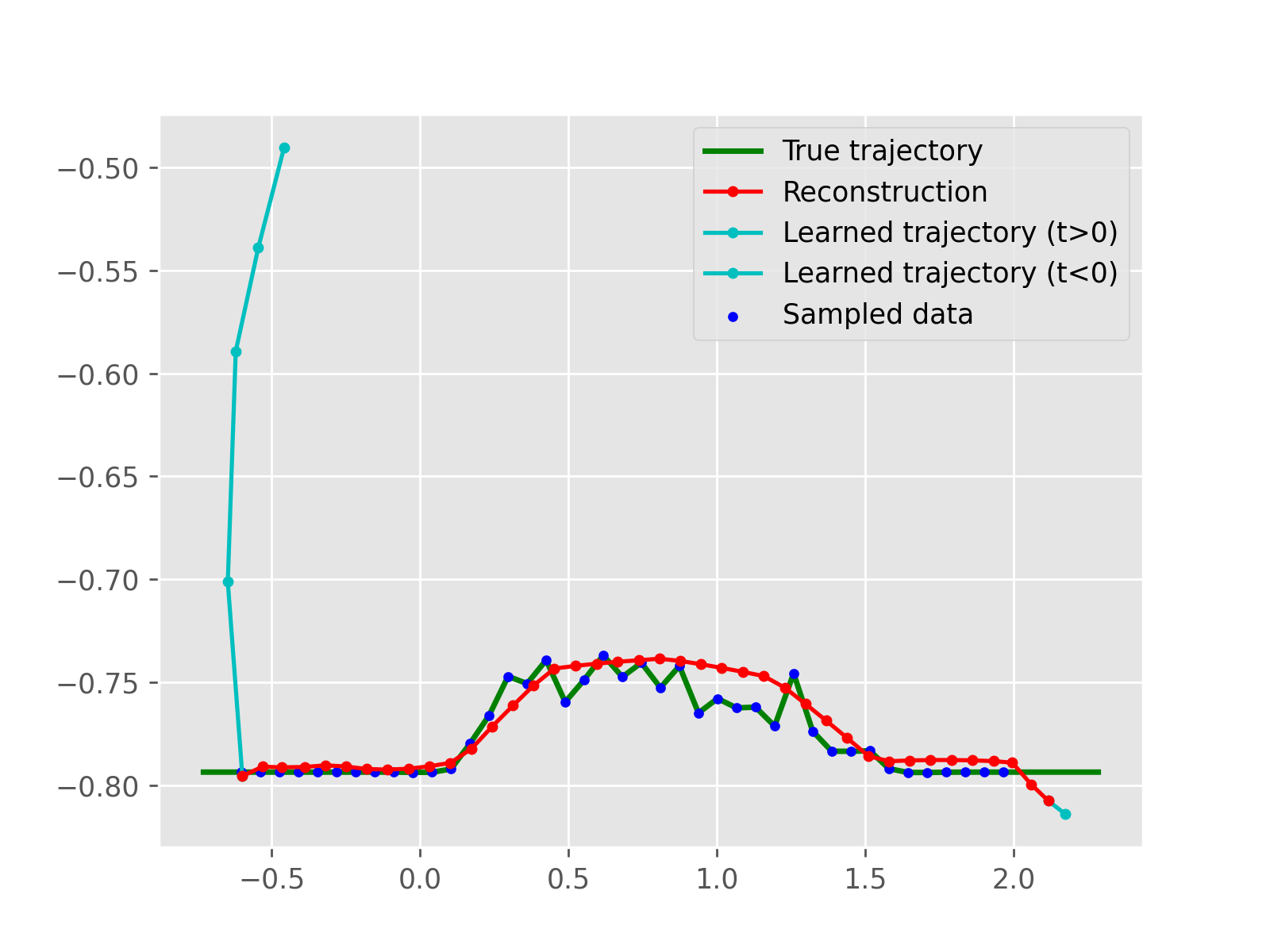}
  \caption{Reconstruction and extrapolation of a single solar power sample using NODE.}
  \label{fig:reconstruction_real}
\end{figure}

In \Cref{fig:reconstruction_real} we can see that the model is able to capture the dynamics of the solar power data in the reconstruction and generate a sinusoidal shape which is representative of the daily power output. However, even for very small values of $t<0$ the extrapolation can be seen to explode. It was not expected that the model was able to extrapolate to nearby days power production, since it was only given single-day sequences, however visually it still shows how the extrapolation can sometimes explode to different order of magnitudes.

\subsection{RMSE}\label{sec:RMSE}
\Cref{tab:rmse} shows the RMSE values for both spiral and solar power as well as increasing complexity of spring data. For the spiral model it can be seen that the NODE VAE model performs better than the baseline with a RMSE of 0.1284 compared to 0.1513. In the appendix \Cref{fig:reconstruction_grid_spiral_baseline} the baseline are visualized and it can be seen to follow the true trajectory with only slight deviations in the beginning of the trajectory.

It is interesting that for the increasing complexity of spring data the RMSE values are found to increase gradually from 0.0174 for Spring (1) to 0.0455 for Spring (1, 2, 3). For the baseline models the RMSE values remain more constant around the 0.04-0.06 range across all complexities of data performing worse than the NODE VAE. It could be investigated further if this trend continues for the NODE VAE, and if the model will begin to struggle if presented for even more complex data than three mixed spring examples. 

For the solar data we find the RMSE for the NODE VAE and the baseline to be very similar with respectively 0.0077 and 0.0068. Both models capture the overall dynamic of the time-series and can reconstruct sinusoidal power curves with the same order of magnitude as real data. None of the models are on the other hand able to capture the intra-day variance in terms of cloud cover and reduced power output. This might be a result of limited data availability and could possibly be improved by including more than only 68 days of data in the analysis.
\begin{table}[h]
\caption{RMSE values for reconstruction for all model runs and baselines.}
\label{tab:rmse}
\begin{tabular}{@{}lll@{}}
\toprule
Experiment       & NODE     & Baseline \\ \midrule
Spiral           & 0.1284 & 0.1513 \\
Spring (1)       & 0.0174 & 0.0647 \\
Spring (2)       & 0.0212 & 0.0582 \\
Spring (3)       & 0.0162 & 0.0471 \\
Spring (1, 2)    & 0.0261 & 0.0601 \\
Spring (1, 3)    & 0.0303 & 0.0525 \\
Spring (2, 3)    & 0.0363 & 0.0510 \\
Spring (1, 2, 3) & 0.0455 & 0.0562 \\
Solar Data       & 0.0077 & 0.0068 \\ \bottomrule
\end{tabular}
\end{table}

\subsection{Training time}\label{sec:training_time}

\Cref{tab:epoch-time} shows the average time per epoch for all performed experiments. The large difference in in the magnitude of average time per epoch is due to the changing amount of samples for training across spirals, springs and solar data. We can see that generally the NODE is faster than the baseline for all spring experiments, while for the spiral example they are relatively close. For the solar data we can see quite a large difference in the average time per epoch with the baseline being faster. This might indicate that for very little data the NODE method does not perform so well in regards to training time, since everything was kept constant between the spring and solar experiments except the number of samples. It is a possibility that the NODE VAE does not take properly advantage of its constant memory property for small amounts of data, but this result is only an indication and something which should be investigated further in future work.

\begin{table}[h]
\caption{Average time pr. epoch for all experiments}
\label{tab:epoch-time}
\begin{tabular}{@{}lll@{}}
\toprule
Experiment       & NODE     & Baseline \\ \midrule
Spiral           & 0.3885 s & 0.3441 s \\
Spring (1)       & 4.9185 s & 5.7555 s \\
Spring (2)       & 4.8733 s & 5.8242 s \\
Spring (3)       & 4.7256 s & 5.4650 s \\
Spring (1, 2)    & 4.7193 s & 7.5129 s \\
Spring (1, 3)    & 4.8526 s & 5.9382 s \\
Spring (2, 3)    & 4.9839 s & 5.9377 s \\
Spring (1, 2, 3) & 4.8614 s & 5.9053 s \\
Solar Data       & 0.1467 s & 0.0533 s \\ \bottomrule
\end{tabular}
\end{table}


\subsection{Latent space}\label{sec:latent_space}
\begin{figure}[h]
  \includegraphics[width=0.45\textwidth]{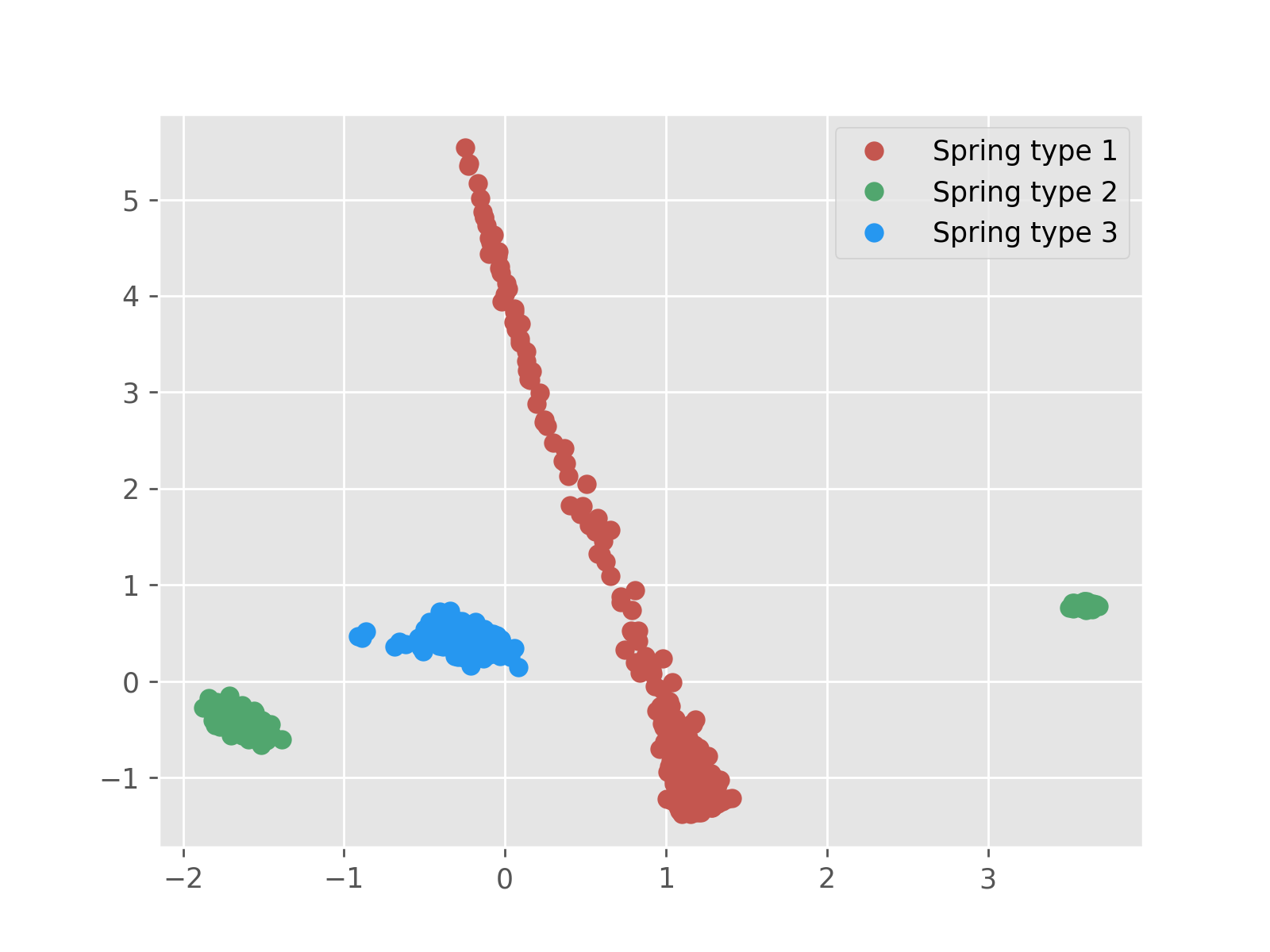}
  \caption{2D projection of $z_0$ sampled from the latent space distribution from the mixed spring 123 model.}
  \label{fig:latent_z0_vis}
\end{figure}

\begin{figure*}[h!]
  \includegraphics[width=1\textwidth]{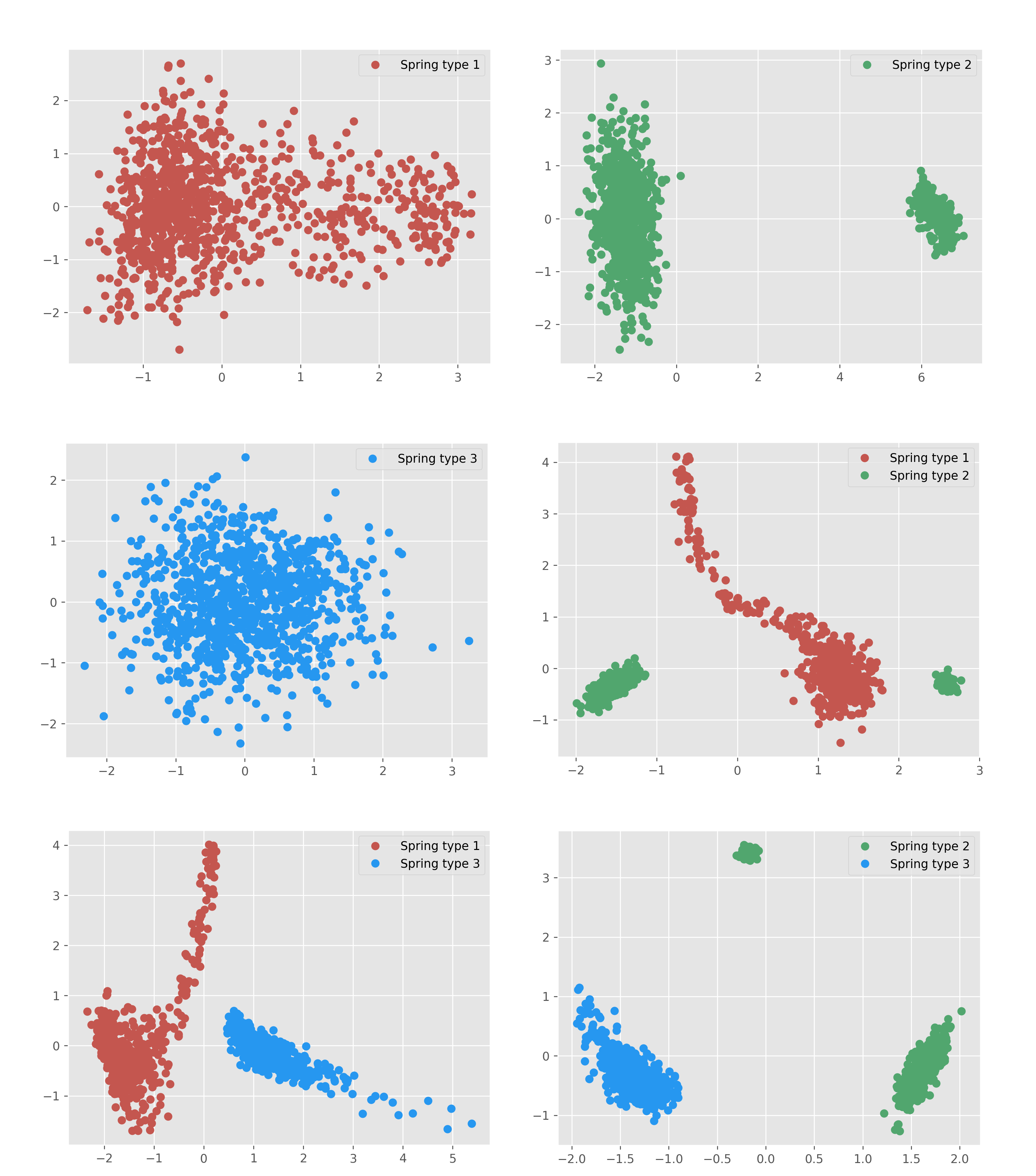}
  \caption{Latent space of the models trained on the different combination spring data.}
  \label{fig:latent_grid}
\end{figure*}

While evaluating the model trained on the Spring (1, 2, 3) data on its test set, the 4-dimensional latent state $\mathbf{z}_{t_0}$ were sampled from the latent space distribution $q(\mathbf{z}_{t_0}|\mathbf{x}_t)$ and have been projected in 2 dimensions using the first two principle components from a Principle Component Analysis (PCA). The two principle components accounts for 85\% of the explained variance, which should give a somewhat reliable interpretation of the projected latent space. The 2D projections are shown in \Cref{fig:latent_z0_vis}. The plot shows a clear distinction in the latent space between the different type of springs used as input in the model. This is a clear indication that the model learns to distinguish different kind of input by sampling from different places in the distribution. Interestingly enough, we see that the spring type 2 data, which is a damped sine wave, appears in two separate clusters in the latent space. This is an indication that the noise we have used when sampling the models most likely have made big enough changes to the input data, that the model thinks spring type 2 are two separate classes of springs. Spring type 1 has a very elongated distribution, most likely indicating that the model has difficulties capturing the dynamics accurately. From the RMSE of the reconstruction, it can also be seen that the spring 1 model has a higher error than the spring 3 model, which could be explained by the model having an easier time reconstructing data that lies closer together in the latent space. Latent space for the other models can be seen in \Cref{fig:latent_grid}.

\section{Conclusion}\label{sec:Con}
To conclude on the work in this paper, we have implemented and built upon the NODE framework presented in \cite{chen2019a} for generative time series modeling using a VAE framework. Several models have been trained across different complexities of input data to test the robustness and performance of the NODE methods. In general the NODE VAE was able to capture the dynamics of the underlying data. We were able to reproduce the results for spirals and found that both forward and backward extrapolation would follow the true trajectory. However, we were not able to train the model to extrapolate for as long times as presented in \cite{chen2019a} (all the way to the center of a spiral). 

When we subjected the NODE VAE to increasing complexity in the data, we found a tendency that the RMSE steadily increased, while for the baseline it remained constant. This might indicate that for even more complex data the NODE VAE will have decreasing performance, which is an interesting hypothesis for the method and should be investigated further in future work. The NODE VAE was also tested in a real-life setting with little data available, and was still able to capture general dynamics in a similar fashion to the baseline. For all generated data sets (spiral and spring) NODE VAE outperformed the baseline but for the real life data, the baseline performed slightly better, in regards to reconstruction error.

Finally, when comparing the training time based on the average time per epoch, it was found that the NODE VAE was generally faster than the baseline for larger numbers of data. For the solar data where only 68 days of data of data was available the baseline outperformed the NODE VAE significantly in regards to training speed. This is most likely because of the limited data, such that the NODE VAE does not take properly advantage of its constant memory property.
\section{Future work} \label{sec:futurework}
Natural extensions to the work performed in this paper would include:
\begin{enumerate}
    \item Investigate further the importance of the input data for the NODE and how the sampling methods affect the results. 
    \begin{itemize}
        \item Perform a similar analysis on samples with irregular timestamps. 
        \item Perform the analysis with Gaussian noise added to the spring data points instead of varying the constants of the three spring equations.
        \item Challenge the NODE model further with more complex data.
    \end{itemize}
    \item Include extrapolation on the baseline to compare the forecasting capabilities with existing methods.
    \item Extend the scope of the research by testing more specific NODE model setups.
    \begin{itemize}
        \item Perform an analysis varying the ODE topology and investigating the impact on model performance.
        \item Test the impact of different ODE solvers.
        \item Test the impact of different neural-net solvers (Adam, SGD).
        \item Test how hyper-parameter tuning would impact the performance.
    \end{itemize}
    
\end{enumerate}


\bibliography{bibliography.bib}

\clearpage
\section*{Appendix}
\begin{figure*}[h!]
  \includegraphics[width=.9\textwidth]{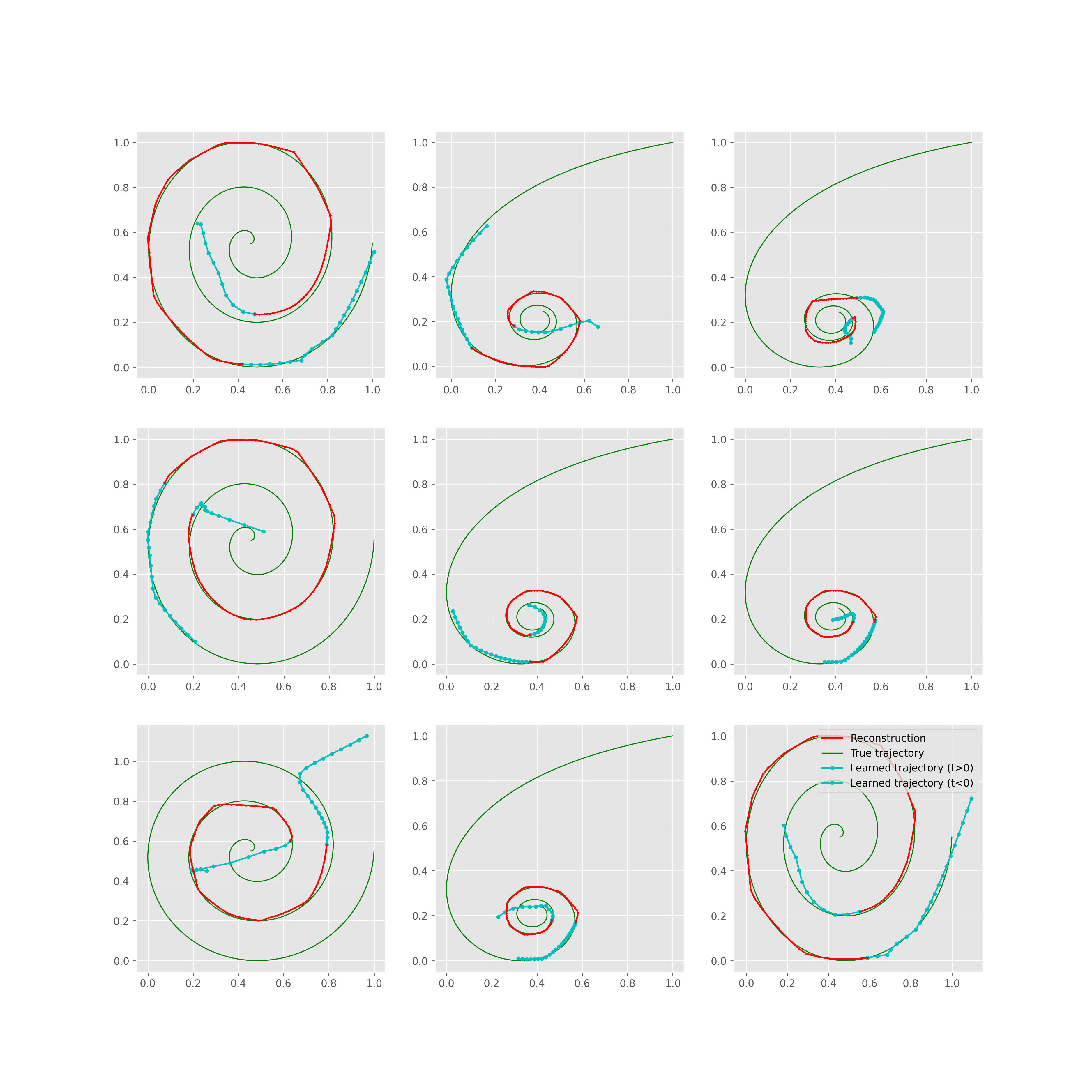}
  \caption{Reconstruction grid for spirals visualizing multiple samples from the NODE.}
  \label{fig:reconstruction_grid_spiral_node}
\end{figure*} 

\begin{figure*}[h]
  \includegraphics[width=1\textwidth]{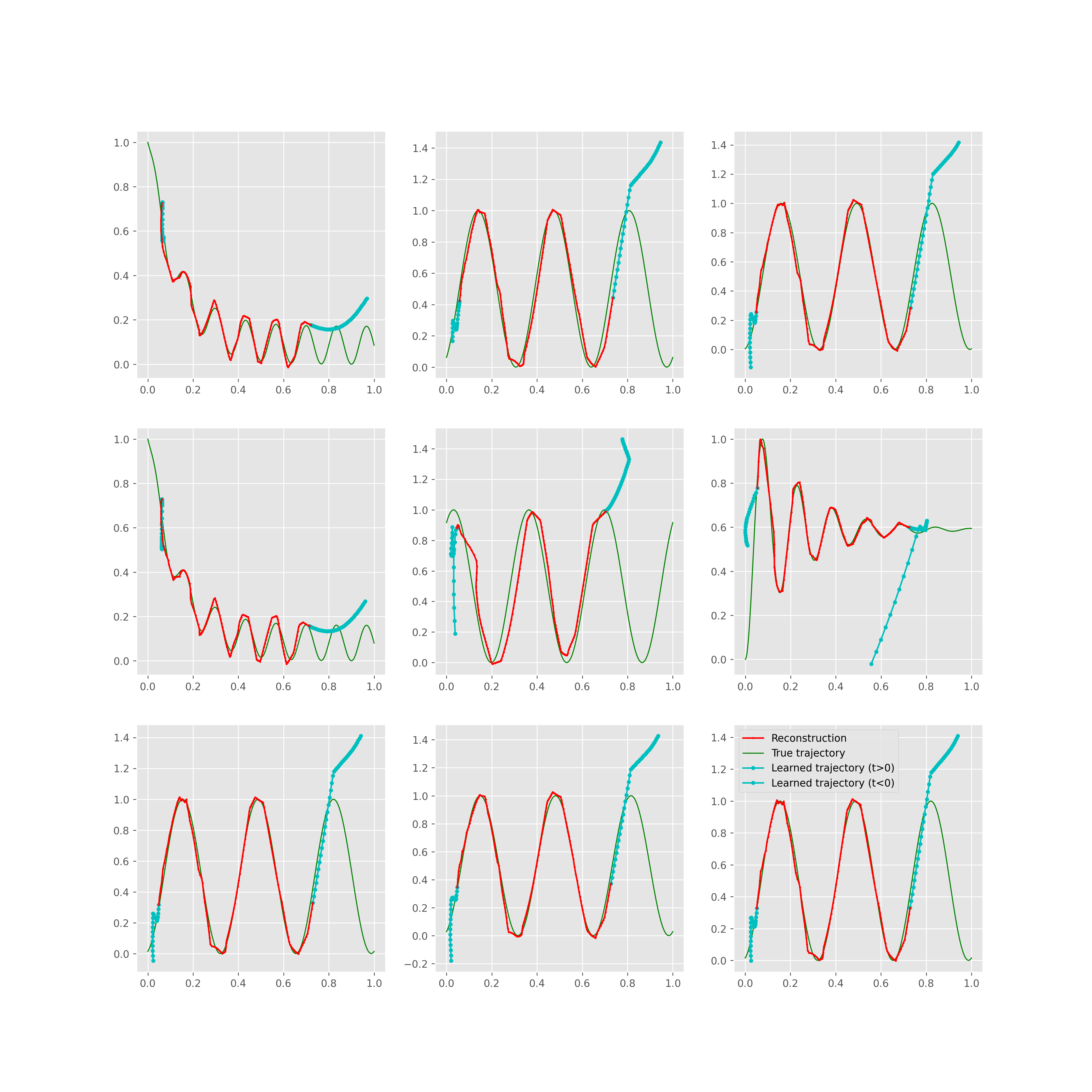}
  \caption{Reconstruction grid for springs with mixed examples 1,2,3 visualizing multiple samples from the NODE.}
  \label{fig:reconstruction_grid_spring_node}
\end{figure*} 
\begin{figure*}[h]
  \includegraphics[width=1\textwidth]{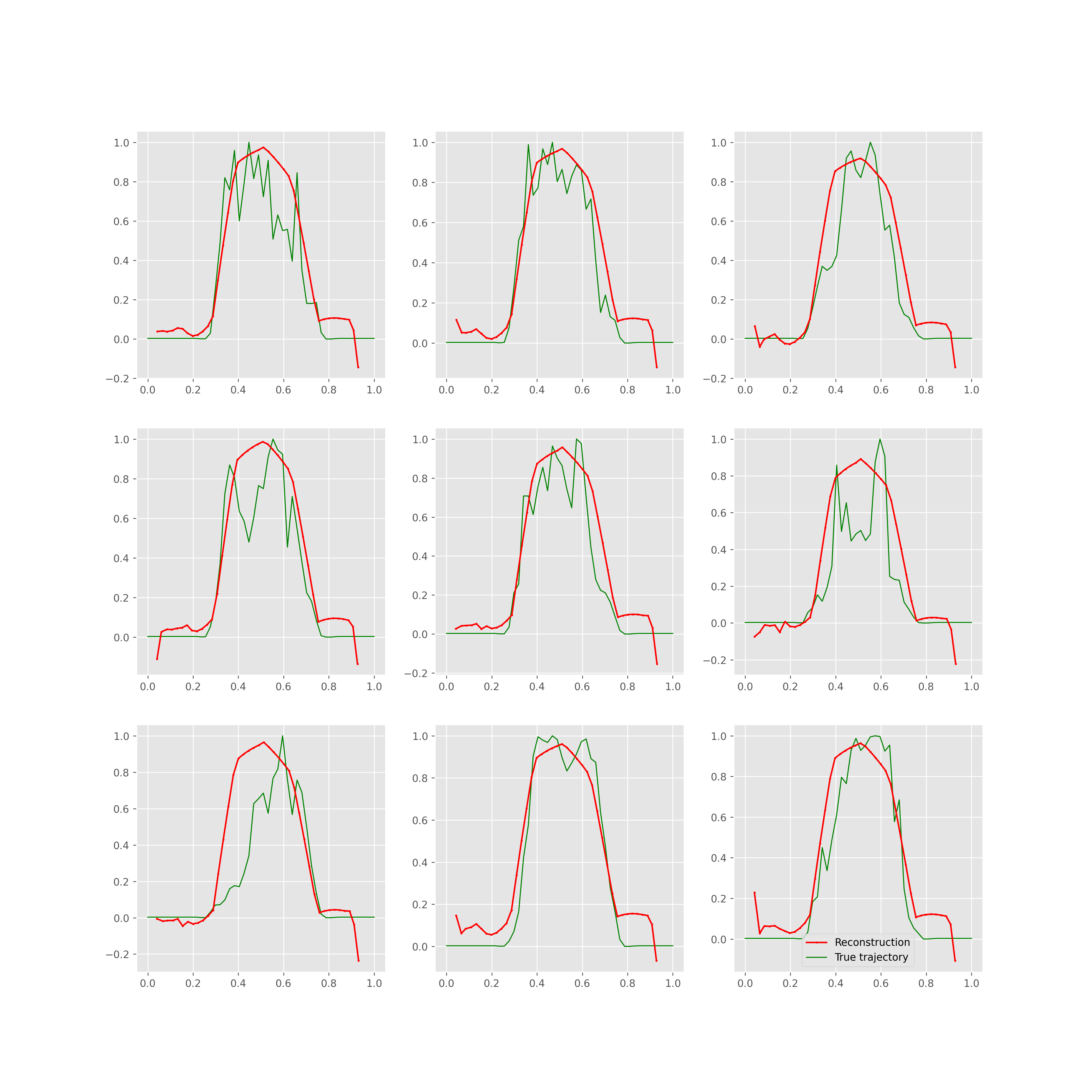}
  \caption{Reconstruction grid for solar power visualizing multiple samples from the NODE.}
  \label{fig:reconstruction_grid_real_node}
\end{figure*} 

\begin{figure*}[h]
  \includegraphics[width=1\textwidth]{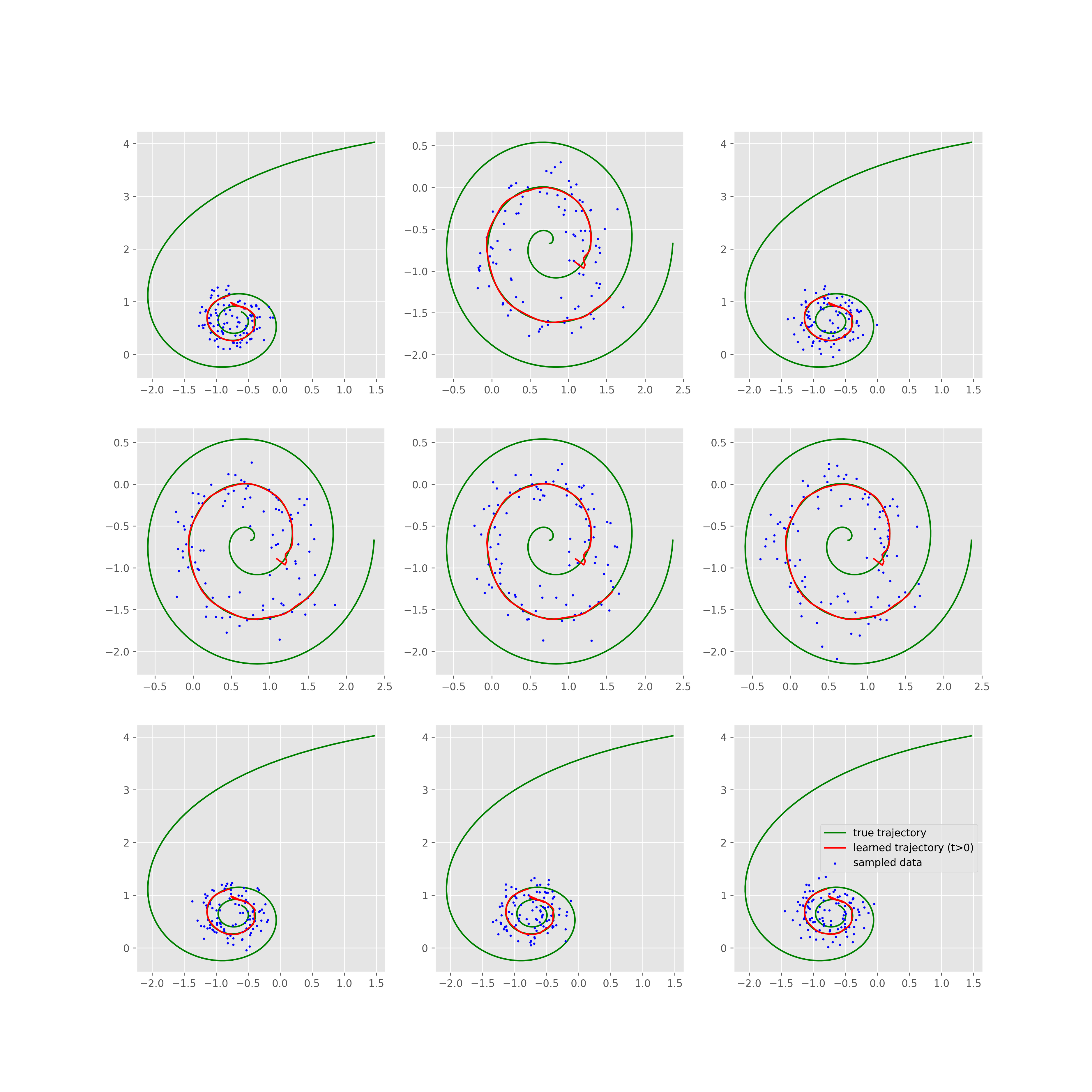}
  \caption{Reconstruction grid for spirals visualizing multiple samples from the baseline.}
  \label{fig:reconstruction_grid_spiral_baseline}
\end{figure*} 

\begin{figure*}[h]
  \includegraphics[width=1\textwidth]{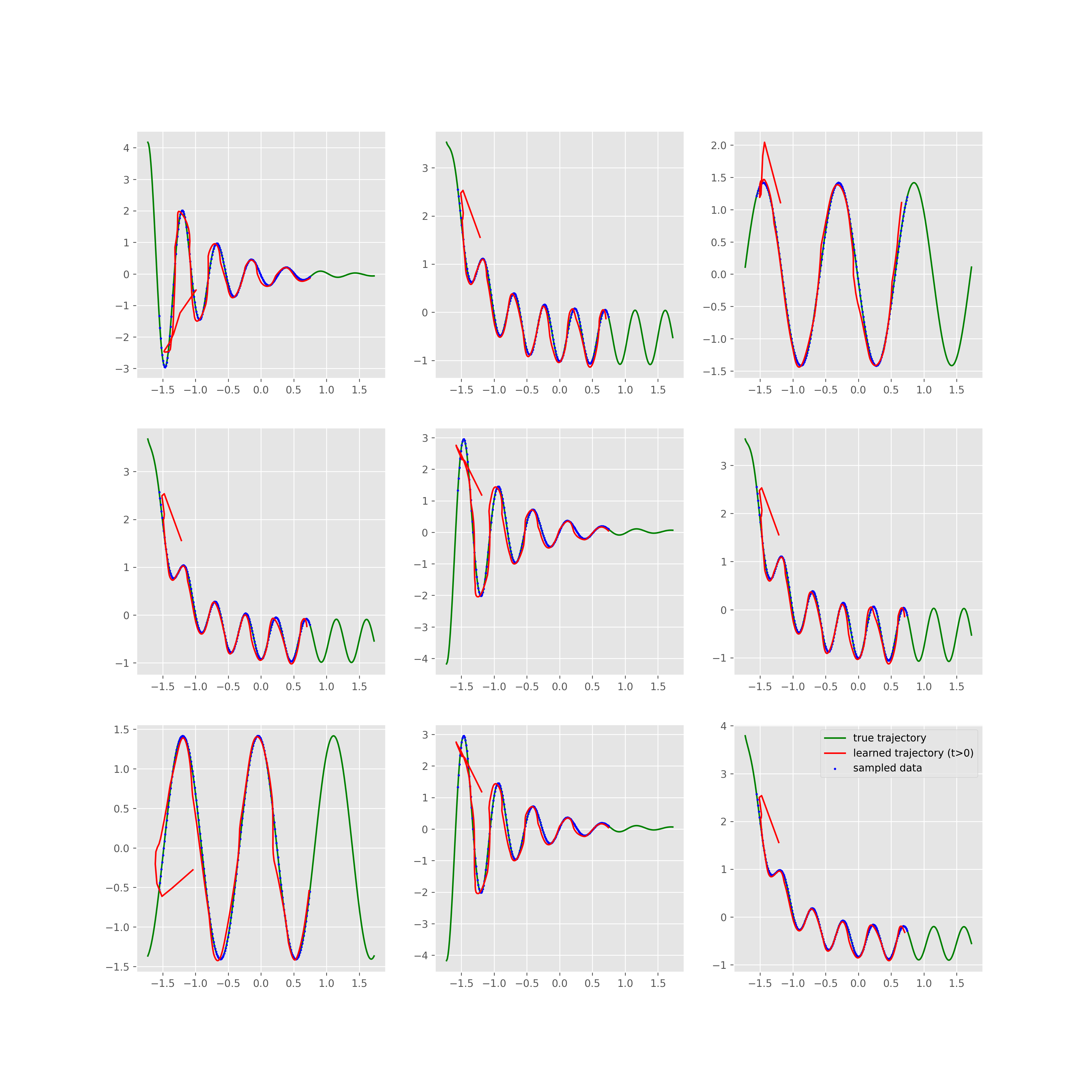}
  \caption{Reconstruction grid for springs with mixed examples 1,2,3 visualizing multiple samples from the baseline.}
  \label{fig:reconstruction_grid_spring_baseline}
\end{figure*} 

\begin{figure*}[h]
  \includegraphics[width=1\textwidth]{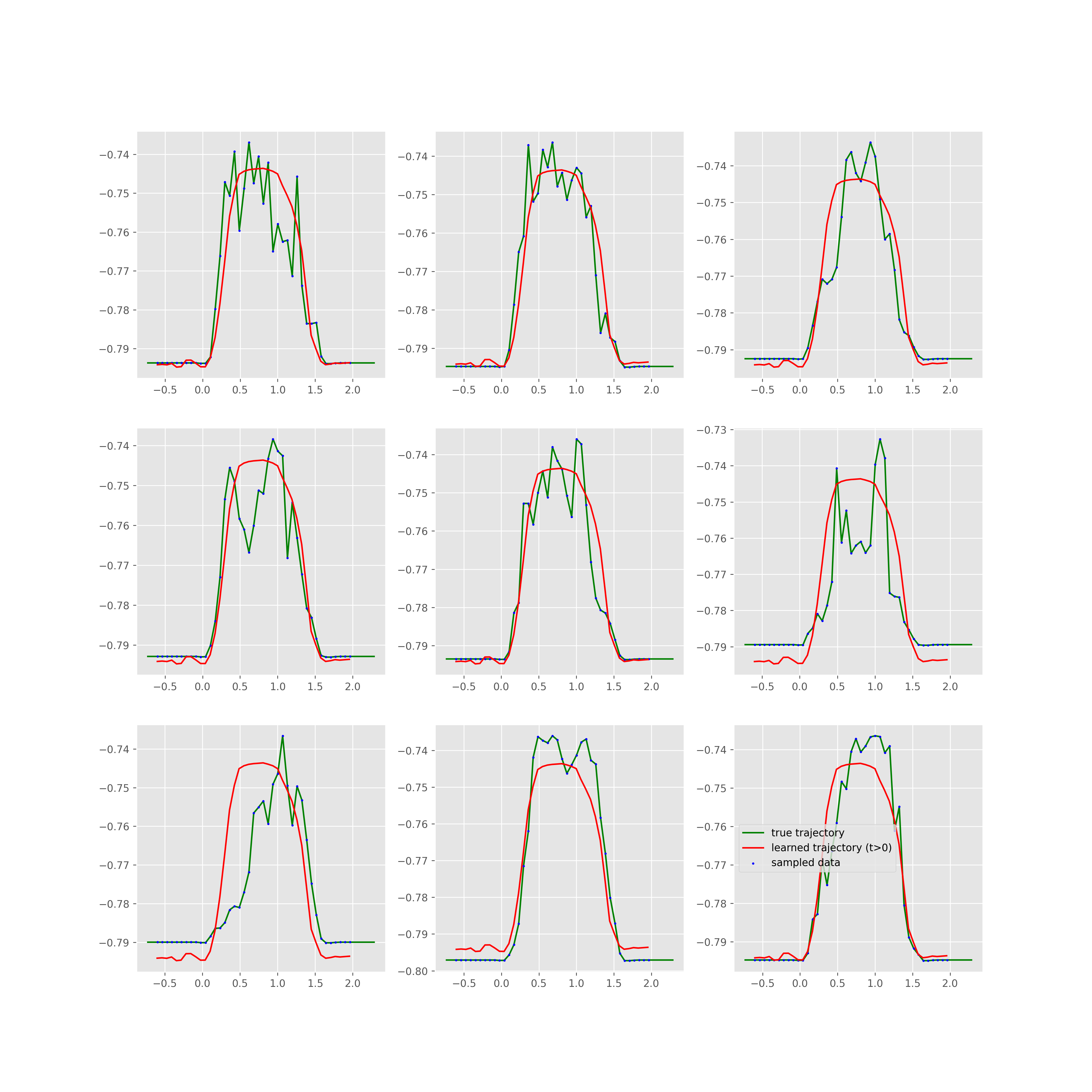}
  \caption{Reconstruction grid for solar power visualizing multiple samples from the baseline.}
  \label{fig:reconstruction_grid_real_baseline}
\end{figure*}

\end{document}